\documentclass[letterpaper, 10 pt, conference]{ieeeconf} 
\IEEEoverridecommandlockouts

\usepackage{graphics} 
\usepackage{epsfig} 
\usepackage{amsmath} 
\usepackage{amssymb}  
\usepackage{algorithm}
\usepackage{algpseudocode}
\usepackage{amssymb}
\usepackage{amsmath}
\usepackage{commath}
\usepackage{mathtools}

\usepackage{hyperref}
\usepackage{import}
\usepackage{paralist}
\usepackage{gensymb}
\usepackage{dsfont}
\usepackage{siunitx}
\usepackage[bottom]{footmisc}
\usepackage{framed}
\usepackage{balance} 
\usepackage{caption}
\usepackage[skins]{tcolorbox}  
\usepackage{subcaption}
\usepackage{bm}
\usepackage{cite} 
\usepackage{balance}

\hypersetup{
    colorlinks=true,
    linkcolor=black,
    citecolor=black,
    filecolor=black,
    urlcolor=black,
}
\usepackage{balance}

\DeclareCaptionFont{mysize}{\fontsize{8}{10}\selectfont}
\usepackage{soul}
\usepackage{hyperref}
\usepackage[percent]{overpic}
\usepackage{xcolor}

\usepackage[subtle,tracking=normal]{savetrees} 
\usepackage[disable]{todonotes}
\def\BibTeX{{\rm B\kern-.05em{\sc i\kern-.025em b}\kern-.08em
    T\kern-.1667em\lower.7ex\hbox{E}\kern-.125emX}}

\newcommand{\multilinecomment}[1]{}

\author{Zeynep Özge Orhan$^{1,2}$, Milad Shafiee$^{2}$, Vincent Juillard$^{1}$,\\ Joel Coelho Oliveira$^{1}$,  Auke Ijspeert$^{2}$, and Mohamed Bouri$^{1}$ 
\thanks{$^{1}$  Zeynep Özge Orhan, Vincent Juillard, Joel Coelho Oliveira, Mohamed Bouri are with the REHAssist group,  Ecole Polytechnique Federale de Lausanne (EPFL). {\tt\footnotesize (e-mail: zeynep.orhan@epfl.ch, vincent.juillard@epfl.ch, joel.coelhooliveira@epfl.ch, mohamed.bouri@epfl.ch)}  }
\thanks{$^{2}$  Zeynep Özge Orhan, Milad Shafiee and Auke Ijspeert are with the BioRobotics Laboratory, Ecole Polytechnique Federale de Lausanne (EPFL). {\tt\footnotesize (e-mail: zeynep.orhan@epfl.ch,  milad.shafiee@epfl.ch, auke.ijspeert@epfl.ch)}  
}
}

\begin{document}

\title{ExoRecovery: Push Recovery with a Lower-Limb Exoskeleton \\ based on Stepping Strategy}

\maketitle

\begin{abstract}
Balance loss is a significant challenge in lower-limb exoskeleton applications, as it can lead to potential falls, thereby impacting user safety and confidence. We introduce a control framework for omnidirectional recovery step planning by online optimization of step duration and position in response to external forces. We map the step duration and position to a human-like foot trajectory, which is then translated into joint trajectories using inverse kinematics. These trajectories are executed via an impedance controller, promoting cooperation between the exoskeleton and the user. 
 Moreover, our framework is based on the concept of the divergent component of motion, also known as the Extrapolated Center of Mass, which has been established as a consistent dynamic for describing human movement. This real-time online optimization framework enhances the adaptability of exoskeleton users under unforeseen forces thereby improving the overall user stability and safety.  To validate the effectiveness of our approach, simulations, and experiments were conducted. Our push recovery experiments employing the exoskeleton in zero-torque mode (without assistance) exhibit an alignment with the exoskeleton's recovery assistance mode, that shows the consistency of the control framework with human intention. To the best of our knowledge, this is the first cooperative push recovery framework for the lower-limb human exoskeleton that relies on the simultaneous adaptation of intra-stride parameters in both frontal and sagittal directions. The proposed control scheme has been validated with human subject experiments.
\end{abstract}

\vspace{-0.1cm}
\section{Introduction}

Lower-limb exoskeletons (LLEs), while promising for assisting those with walking impairments, face significant challenges in maintaining balance and stability, particularly in real-world scenarios with various external disturbances~\cite{Monaco2017, Hamza2020, Inkol2022}. Balance is a generic term describing the body posture dynamics to prevent falling and it is a crucial ability to ensure upright standing \cite{Winter1995, Tokur2020}. Researchers revealed three elementary strategies of human balance recovery such as ankle and hip strategies, and stepping strategy with variable step duration \cite{Horak1986, Kuo1993, MacKinnon1993,  Winter1995}. When the disturbances are too large to handle with the aforementioned strategies, they are tackled with a stepping strategy.

When individuals experience balance perturbations, executing a well-coordinated step can help restore equilibrium and prevent falls. By adjusting the position of the feet, the stepping strategy allows individuals to shift their center of mass (CoM), stabilize their posture, and counteract external disturbances. The implementation of an effective stepping strategy with a suitable step length and step duration is particularly important in LLEs, as it enables users to regain balance and navigate safely in challenging environments.

\begin{figure}[t]
    \centering
    \includegraphics[width=0.52\linewidth]{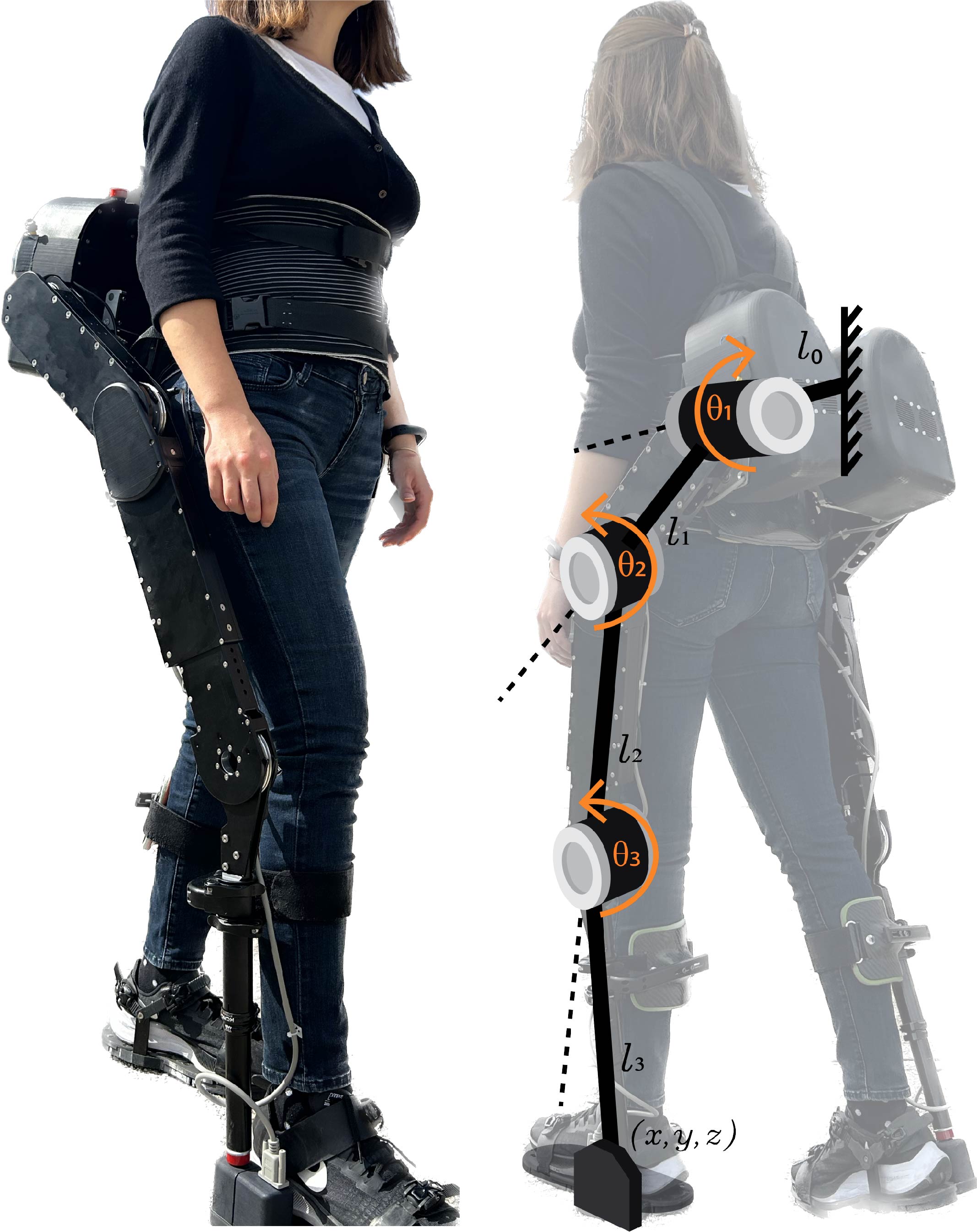}\\
    \vspace{-0.12cm}
    \caption{The 2nd prototype of autonomyo lower-limb exoskeleton. A simplified kinematics model of the autonomyo exoskeleton is illustrated.}
    \label{fig:exo}
\end{figure}

Although the majority of the studies focused on gait assistance with LLEs, balance control in standing and walking is crucial for the control of an LLE for posture and gait assistance \cite{Ringhof2019, Qiu2019}. Recently, the balance strategies of humans also inspired researchers in the field of LLEs to assist the balance of users \cite{Vlutters2018, Qiu2019}.

Most of the effort in the standing balance assistance is taking advantage of actuated ankle joints \cite{Ugurlu2016, Emmens2018, Zha2019, Yin2022, Bayon2022, Beck2023}. There are other studies that investigated ankle, hip, and combined strategy in terms of subject-specific stability limits, slip prevention, importance of hip abd./add. for weight shift and lateral foot placement \cite{Rajasekaran2015, Monaco2017, Wang2015}. 

Duburcq et al. \cite{Duburcq2022} pioneered a push recovery controller based on reinforcement learning, marking a substantial advancement. They successfully demonstrated reactive push recovery with a humanoid robot, utilizing deep reinforcement learning. Notably, this study did not involve direct interaction between the exoskeleton and the user, and the exoskeleton functioned primarily as a humanoid robot.

In the study by Zhang et al. \cite{Zhang2021}, they proposed a method based on the capture point (CP) concept to enhance the balance restoration capabilities of LLEs under significant interference conditions. However, this study did not address step duration optimization, and its focus was exclusively on the sagittal plane.
Similarly, in \cite{Guo2019}, an xCoM-based balance assistance strategy in the sagittal plane is proposed for disturbances under forward and backward directions. Vallery et al. \cite{Vallery2012}, proposed another xCoM-based balance controller that provides calculated feed-forward trajectories. The controller is triggered in case of balance loss and is used to keep users in an upright standing position for a stationary exoskeleton. In \cite{Hua2022}, a balance assistance strategy in the sagittal plane is suggested based on the zero-moment point (ZMP) model-based method. The ZMP has been used to generate the trajectory of movement during the stance phase and assistive torques are designed based on the minimization of modulating of virtual potential energy.

It's worth noting that the CP, Divergent Component of Motion (DCM) and the xCoM share the same definition. The terms CP and DCM are commonly used within the robotics community, while xCoM is more frequently employed in the biomechanics community.
Although CP and ZMP-based methods are widely used in the literature on humanoids \cite{shafiee2017robust,jeong2019robust,kim2023model,shafiee2019online,khadiv2020walking} and also for walking pattern generation for exoskeletons~\cite{Wang2015, Zhang2017, huang2022divergent}, it has not been investigated in detail for step position and duration adaptation of exoskeletons in sagittal and frontal planes. Furthermore, these methods were tested in simulations, leaving room for further hardware validation \cite{Peng2017, Inkol2020, Sergey2016, Huynh2016, Jatsun2017}. 
Farkhatdinov et al. \cite{Farkhatdinov2019} introduced a push recovery mechanism for the human-exoskeleton. However, this approach primarily focused on applying assistive torque without model-based optimization or consideration of the system's dynamics.

As the exoskeleton user actively participates in the recovery step, it is crucial to override the user's behaviors to provide assistance. Instead, the exoskeleton and the user symbiotically move toward the human desired recovery pose. To the best of our knowledge, currently, there is no robotic exoskeleton that can support adaptive step position and duration for balance recovery in sagittal and frontal planes that are collaboratively interacting with exoskeleton users.

The present work describes an experimental study on exoskeleton-assisted recovery stepping together with its theoretical background for balance recovery in case of perturbation in healthy individuals. 
The introduced framework offers targeted joint trajectories to facilitate adjustments in stepping location, encompassing hip abd./add. for step-width adaptation, as well as knee and hip flex./ext. for step-length adaptation. In particular, the contributions of this paper are the following:
\vspace{-0.55cm}
\begin{itemize}
    \item We have developed a user-cooperative omnidirectional recovery stepping control strategy in case of a balance loss. The suggested strategy is implemented and verified through simulations and conducted in-lab experimental evaluations with human subjects.
    \item We propose a framework to detect a possible balance loss under severe perturbations based on divergent component of motion dynamics. 
    \item We have implemented a bio-inspired recovery step trajectory based on human foot position during gait.
\end{itemize}

\vspace{-0.5cm}
\section{Methods}

\subsection{Linear Inverted Pendulum Model (LIPM)}

\begin{figure}[!t]
    \centering
    \includegraphics[width=0.5\linewidth]{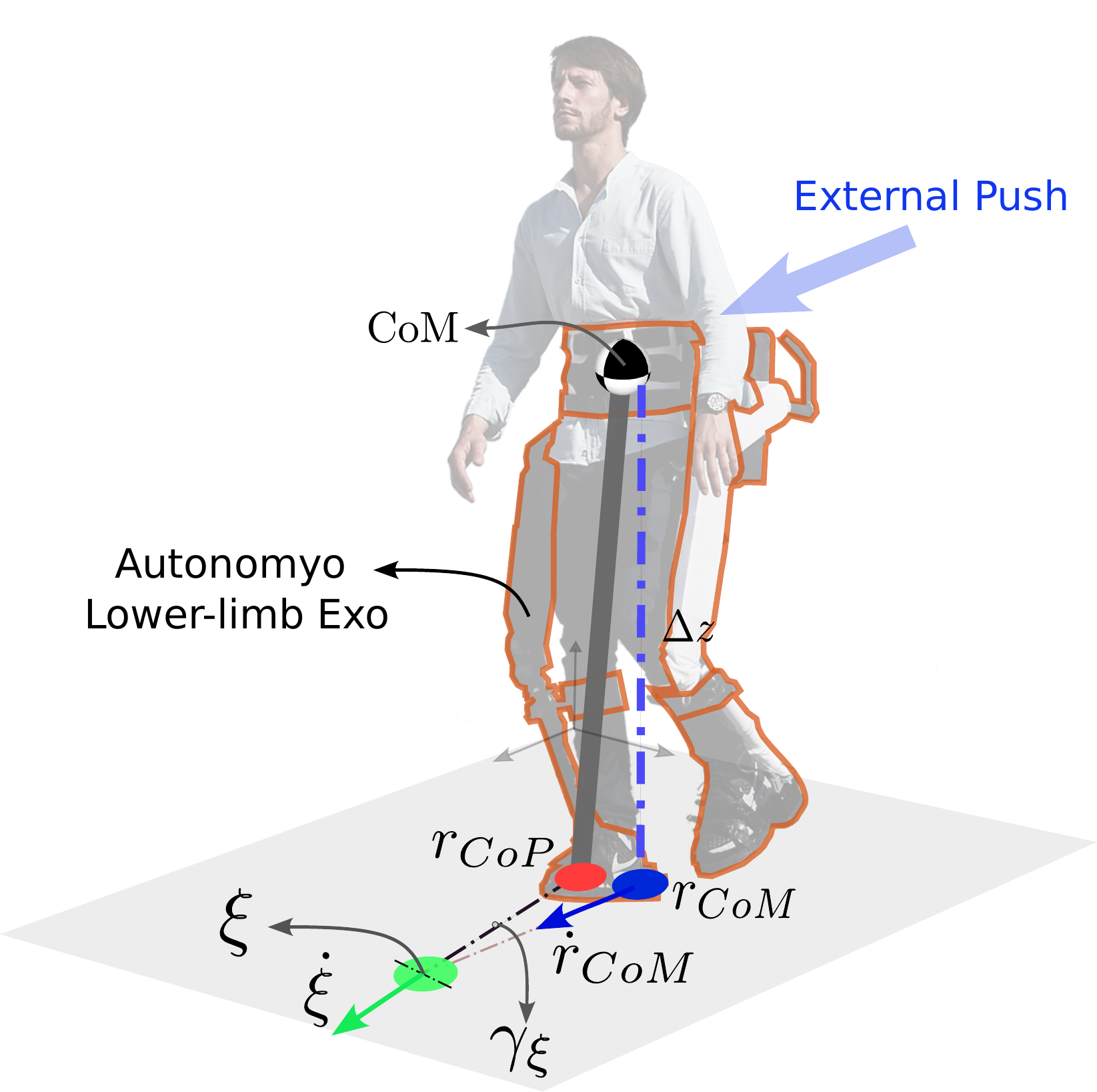}\\
    \caption{DCM, CoM and CoP Points correlations for Centroidal Dynamics}
    \label{fig:dcmDyanamics}
\end{figure}

The LIPM has been widely utilized to describe the dynamics of the CoM for bipedal locomotion~\cite{kajita2003biped}. The LIPM assumes a constant rate of change of centroidal angular momentum and movement of the CoM height within a plane. In~\cite{Beck2023}, it has been suggested that for symbiotic human-exoskeleton balance control, CoM kinematics-based feedback could be beneficial to precede physiological responses. Also, these assumptions are suitable for planning humanoid robot locomotion, as research on human walking indicates minimal variations in centroidal angular momentum and CoM height\cite{herr2008angular}. Based on the assumptions, the equations of motion for the LIPM can be derived as follows:
\begin{equation}
\vspace{-0.25cm}
 \mathbf{ \ddot x}= \omega^{2}( \mathbf{x}-\mathbf{cop}) 
 \label{eq:1}
\vspace{-0.02cm}
\end{equation}
in which $\mathbf{ x}=[ x_{com}, y_{com}]^T$ is the horizontal position of the CoM,  $\omega_{0}=\sqrt{\frac{g}{\Delta z}}$ is the natural frequency of the LIPM,  and $\mathbf{cop}=[ cop_{x}, cop_{y}]^T$  is the horizontal position of the center of pressure (CoP). 

\vspace{-0.15cm}
\subsection{ Divergent Component of Motion}
In this section, we provide an overview of the DCM concept's background. The dynamics of the CoM, as modeled by the LIPM, can be split into stable and unstable components. The unstable component is referred to as the DCM and is defined as follows:
\vspace{-0.2cm}
\begin{equation}
\boldsymbol{\xi}=\mathbf{ x}+\frac{\mathbf{\dot x}}{{\omega}} 
  \label{eq:2}
\vspace{-0.1cm}
\end{equation}

Throughout this study, DCM is represented as $\xi$ as the notation in \cite{takenaka2009real}. From (\ref{eq:2}), the CoM dynamics is given by:
\vspace{-0.2cm}
\begin{equation}
\mathbf{\dot x}={{\omega}} (\boldsymbol{\xi}-\mathbf{ x} )
  \label{eq:3}
  \vspace{-0.1cm}
\end{equation}
By differentiating (\ref{eq:2}) and substituting (\ref{eq:1}), the DCM dynamics is expressed as :
\vspace{-0.2cm}
\begin{equation}
\dot{\boldsymbol{ \xi}}={{\omega}}(\boldsymbol{\xi}-\mathbf{cop})
  \label{eq:dcmDynamics}
\vspace{-0.2cm}
\end{equation}
Fig. (\ref{fig:dcmDyanamics}) illustrates the relationship between DCM dynamics, CoM, and the CoP. By re-arranging DCM dynamics \eqref{eq:dcmDynamics}, the following ordinary differential equation (ODE) holds:
\vspace{-0.2cm}
\begin{equation}
\dot{ \boldsymbol \xi} - \omega{\boldsymbol \xi}=-{\omega} \hspace{0.072 cm}\mathbf{cop}_0
\label{eq:DCMode}
\vspace{-0.2cm}
\end{equation}
The solution to \eqref{eq:DCMode} writes:
\vspace{-0.2cm}
\begin{equation}
\boldsymbol\xi(t)=e^{ \int {\omega} dt} \left[  \int \left( -   {\mathbf{cop}_0} \hspace{0.072 cm} \omega\right)  e^{\int -{\omega} dt}  dt+ \mathbf{C}  \right],
 \label{eq:integralDCM}
\vspace{-0.2cm}
\end{equation}

where $C\in \mathbb{R}^2$  is the vector of unknown coefficients that can be found by imposing the boundary conditions. Therefore, we can find these coefficients by solving the problem \eqref{eq:integralDCM} either as an initial value problem, namely
 \vspace{-0.2cm}
 \begin{equation}
 \label{eq:initial_value_problem}
\boldsymbol \xi(0)= \boldsymbol \xi_0=\mathbf{cop}_0+\mathbf{C}_0, 
\vspace{-0.1cm}
\end{equation}
or as a final value problem:
\vspace{-0.2cm}
\begin{equation}
\label{eq:final_value_problem}
\boldsymbol\xi(T)=\boldsymbol\xi_T=\mathbf{cop_0}+ \mathbf{C_f}  \hspace{0.032 cm} e^{\omega{T} }.
\vspace{-0.1cm}
\end{equation}

Therefore, by solving the equation (\ref{eq:dcmDynamics}) as an initial value problem, we arrive at the following equation that represents the time evolution of the DCM:
\vspace{-0.2cm}
\begin{equation}
\boldsymbol{\xi}=(\boldsymbol{\xi}_{0}-{\mathbf{{cop_0}}})\hspace{0.05cm}{{\exp}}({{\omega}}{{t}})+\mathbf{{cop_0}}
  \label{eq:6}
  \vspace{-0.1cm}
\end{equation}

We also can solve the CoM dynamics (\ref{eq:3}) by treating it as an initial value problem: 
\vspace{-0.2cm}
\begin{equation}
\mathbf{x}=(\mathbf{{x}_{0}}-\boldsymbol{\xi}_{0})\hspace{0.05cm}{{\exp}}({{-\omega}}{{t}})+\boldsymbol{\xi}_{0}
  \label{eq:5}
  \vspace{-0.1cm}
 \end{equation}
As evident from the above equation, the CoM exhibits stable dynamics, with the exponential term being negative. However, the DCM exhibits unstable dynamics, characterized by a positive exponential term. This indicates that the difference between $\xi_0$ and ${cop_0}$ increases exponentially over time. Therefore, a prerequisite for ensuring the stability of the CoM trajectory is that the DCM trajectory remains stable.
As stated in \cite{hof2008extrapolated}, the concept of DCM provides a relatively straightforward approach for formulating stability requirements in walking. A simple rule proves effective in ensuring walking stability: when placing the foot, position the Center of Pressure (CoP) at a certain distance behind and outward from the DCM at the moment of foot contact. This distance between the CoP and the DCM during foot placement is referred to as the DCM offset, and minimizing this distance is crucial for maintaining viable states.

\vspace{-0.2cm}
\subsection{Step Adaptation Controller} 
\label{sec:stepAdaptation}
This section presents the step adaption mechanism based on DCM dynamics~\cite{khadiv2020walking,shafiee2019online}.
More precisely, we present below a step adjustment strategy that optimizes the next step position and timing based on the measured DCM. 

To find a DCM trajectory that satisfies both the initial and the final condition problems, the coefficient $C_0$ must equal $C_f$. Thus, by combining \eqref{eq:initial_value_problem} and \eqref{eq:final_value_problem}, one has:
\vspace{-0.2cm}
\begin{equation}
\boldsymbol \xi_0-\mathbf{cop}_0=\left( \boldsymbol \xi_T-\mathbf{cop}_0\right)  e^{-\omega {T}}.
 \label{eq:constCoeef} 
\end{equation}
Now by defining $\sigma=e^{\omega {T}}$ we obtain : 
\vspace{-0.2cm}
\begin{equation}
\boldsymbol \xi_T+ \mathbf{cop}_0 (-1 +   \sigma)- \boldsymbol \xi_0  \sigma=0.
 \label{eq:DCMConstraint0} 
 \vspace{-0.1cm}
\end{equation}
Let $\mathbf{cop}_T$ represent the CoP position at the start of the next step, and $\boldsymbol{\gamma_T} = \boldsymbol{\xi_T} - \mathbf{cop_T}$ denote the DCM offset for the next step (i.e, the end of this step). Therefore, straightforward calculations lead to:
\vspace{-0.2cm}
\begin{equation}
 \boldsymbol{\gamma_T}+\mathbf{cop}_T+ \left(\mathbf{cop}_0- \boldsymbol \xi_0  \right)\sigma=\mathbf{cop}_0 .
 \label{eq:DCMConstraint1} 
 \vspace{-0.1cm}
\end{equation}

The step adjustment problem can be formalized as a constrained optimization problem, wherein the search variables consist of $\gamma_T$, $\mathbf{cop}_T$, and $\sigma$, and the cost function is appropriately defined in a quadratic manner. It's worth noting that the desired final DCM position and step timing are dependent on $\gamma_T$ and $\sigma$, respectively. Additionally, $\mathbf{cop}_T$ is assumed to be located at the center of the foot at the start of the next step. Therefore, we can treat this position as the target for the upcoming footstep placement.
The selected cost function aims to minimize the deviation of the desired gait values from the nominal ones:
\vspace{-0.2cm}
\begin{equation}
\begin{split}
J &=  \alpha_1  \norm{\mathbf{cop}_{T} - {\mathbf{cop}}_{T,nom}}^2 + \alpha_2  \norm{\boldsymbol\gamma_{T} - \boldsymbol\gamma_{nom}  }^2 \\
&+ \alpha_3  |\sigma -  e^{\omega{T_{nom}}  } |^2,
 \label{eq:costStep} 
 \end{split}
 \vspace{-0.5cm}
\end{equation}
where $\alpha_1$, $\alpha_2$, $\alpha_3$ are positive numbers and the next ZMP position $\mathbf{cop}_{T,nom}$, step duration $T_{nom}$ and next DCM offset $\gamma_{nom}$ are the desired values.
\par
We also present the following set of inequality constraints:
\vspace{-0.2cm}
\begin{equation}
\begin{bmatrix}
 I_{2} &  0_{2 \times 1}   &0_{2}  \\
  -I_{2} &  0_{2 \times 1}   &0_{2}  \\
  0_{1 \times 2} & I_1  & 0_{1 \times 2}  \\
  0_{1 \times 2} & -I_1  & 0_{1 \times 2}  
\end{bmatrix}
\begin{bmatrix}
 \mathbf{cop}_{T}  \\
 \sigma  \\
 \boldsymbol \gamma_{T}  
\end{bmatrix}
\le
\begin{bmatrix}
 \mathbf{cop}_{T,max}\\
  - \mathbf{cop}_{T,min} \\
  \sigma_{max} \\ 
     -\sigma_{min}\\
\end{bmatrix},
\vspace{-0.3cm}
\end{equation}
Here, $\mathbf{cop}_{T,max}$ and $\mathbf{cop}_{T,min}$ are in $\mathbb{R}^2$, while $\sigma_{max}$ and $\sigma_{min}$ belong to $\mathbb{R}$. These inequality constraints are established considering the constraints imposed by leg kinematics on the maximum step length and by the maximum achievable velocity on the minimum step duration.
Lastly, the relationship described in \eqref{eq:DCMConstraint1} is considered as an equality constraint.
Due to the quadratic and linear dependence of the cost function and constraints on the unknown variables, the entire framework can be formulated as a Quadratic Programming (QP) problem.
At each control cycle, the QP problem is solved by substituting $\xi_0$ with the current DCM position and dynamically shrinking the single support duration as the robot executes the step. It is worth noting that during push recovery with a human-exoskeleton, the controller attempts to minimize the DCM offset. However, at the end of the recovery step, the capturability constraint for stopping movement, which mandates a DCM offset of zero, is managed with the use of the exoskeleton, employing an ankle strategy. This is consistent with the concept of an assisting mode where humans and robots collaborate to maintain balance.

\vspace{-0.2cm}
\subsection{Control Strategy Implementation}
\label{sec:ControlStrategy}
\vspace{-0.1cm}
\subsubsection{autonomyo exoskeleton}
The LLE autonomyo is developed to partially assist people who have walking impairments due to neuromuscular deficits \cite{Ortlieb2017}. The second prototype of the autonomyo exoskeleton (autonomyo v2) has 6 degrees of freedom (DoF), 3 active and 3 passive per leg, as in the first design. The 3 active DoF are on the hip (abd./add. and flex./ext.) and knee (flex./ext.) joints as shown in Fig. \ref{fig:exo}. The remaining passive DoF are at the ankle joint (eversion/inversion, dorsiflex./plantar flex., and abd./add.). 

For hip and knee flex./ext. actuation, each unit consists of a brushless motor (BP4, Faulhaber AG, Switzerland) and a corresponding gearbox (42GPT, Faulhaber AG, Switzerland) with a 108:1 transmission ratio together with an integrated torque sensor at the actuator side. An additional cable transmission (2.6:1) is utilized for hip and knee flex./ext. actuation, while a ball screw transmission is used for hip abd./add.. 

The actuators, batteries, and electronics of the system are mainly placed in the back modules. The exoskeleton has three interfaces at the foot, the shank, and the trunk for the physical connection to the user. The weight of the device is about 20 kg including the batteries. The size of the lower body segments is adjustable according to user-specific measurements. The controllers are implemented on the embedded computer of autonomyo v2 (BeagleBone Black, Texas Instruments, USA). Wireless communication with autonomyo v2 is established through a Wi-Fi module.

\subsubsection{Center of Mass Estimation}
The CoM position is estimated based on the trunk roll and pitch angles that are reconstructed through the accelerometer and gyroscope data collected through the MPU6050 IMU module on the back module of the exoskeleton. The CoM is assumed at a constant height that has been in the same place as the sensor placement. The exoskeleton is considered as a rigid leg with a fixed length during the stance pose and for small angles.

\subsubsection{Balance Loss Detection}
For effective balance assistance, real-time assessment of balance and timely detection of upcoming balance loss are paramount. Posturography, a reliable method for objectively quantifying postural sway and balance control, traditionally relies on force plate measurements to assess ground reaction forces~\cite{Hasselkus1975}. A key metric in this assessment is the sway area, the $95\%$ confidence ellipse around the mean postural sway in both anteroposterior and mediolateral directions~\cite{Quijoux2021}.

Since the postural sway is based on the movement of the body's CoM \cite{Maudsley-Barton2020}, to set the limits for detecting these perturbations, we were inspired by the elliptical shape often used to represent the sway area. This choice allows us to effectively define boundaries within which DCM dynamics can be considered normal or indicative of a perturbation event. 

\begin{figure}[t]
    \centering
    \includegraphics[width=1\linewidth]{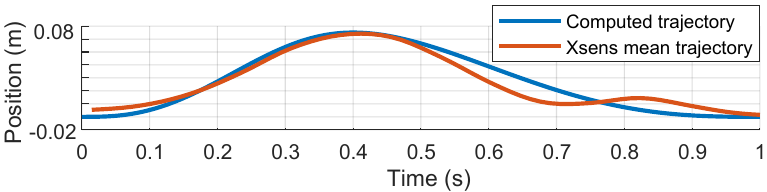}\\
    \vspace{-0.2cm}
    \caption{Comparison of foot trajectory data collected during walking to the designed foot trajectory}
    \label{fig:trajectoryDesign}
\end{figure}

\begin{figure}[b]
    \centering
    \includegraphics[width=0.90\linewidth]{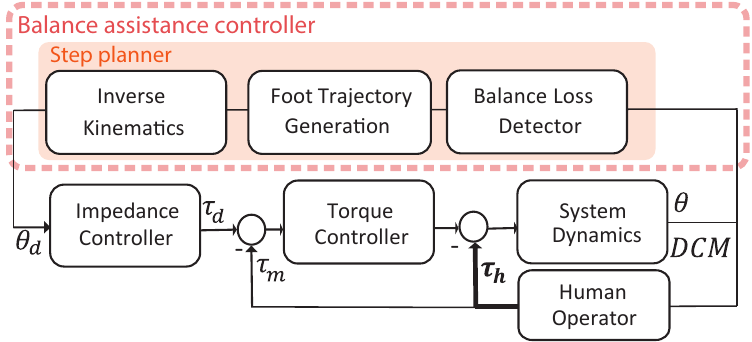}
\caption{The overview of the controller scheme. The step planner is activated when a recovery step is required based on the balance loss detector.}
\label{fig:controldiagram}
\end{figure}
\vspace{-0.02cm}

\subsubsection{Inverse Kinematics of autonomyo}
Each side of the exoskeleton is modeled as a three-degree-of-freedom robotic arm considering the foot is an end-effector and the trunk is the ground as shown in Fig. \ref{fig:exo}. The desired foot trajectory is denoted as ${x,y,z}$, where $l_0$ and $l_1$ are the distances between the middle of the trunk connection to the center of rotation of the hip abd./add. joint, between the center of rotations of the hip abd./add. and hip flex./ext. joints. The thigh length is shown as $l_2$, and $l_3$ is the shank length. 

A geometric approach is followed to obtain an analytical equation for the inverse kinematics. Since the desired trajectory for the foot is designed ${x,y,z}$, through the inverse kinematics hip abd./add., hip flex./ext., and knee flex./ext. joint trajectories are generated by using Eqn.~\ref{eq:IK}.

\vspace{-0.6cm}
\begin{equation}
    \begin{split}
        \theta_1 & = \frac{\pi}{2} + \text{arctan}(\frac{y}{z})  - \text{arctan}(\frac{\sqrt{r^2-d^2}}{d})\\
        \theta_3 & = \text{arctan}(\frac{-D}{\sqrt{1-D^2}})\\
        \theta_2 & = \text{arctan}(\frac{\sqrt{r^2-d^2}}{x}) - \text{arctan}(\frac{{l_2}+{l_3}\cos\theta_3}{{l_3}\sin\theta_3})
    \end{split}
    \label{eq:IK}
\end{equation}
where $r^2 = y^2 + z^2$ and $D = \frac{r^2-{l_1}^2+x^2-{l_2}^2-{l_3}^2}{2{l_2}{l_3}}$.The $\theta_1$, $\theta_2$, and $\theta_3$ are the hip abd./add., hip flex./ext., and knee flex./ext. joint angles, respectively. 

\subsubsection{Foot Trajectory Design}

Once the optimization problem described in Sec.\ref{sec:stepAdaptation} is solved, the desired foot placement is obtained. From the initial point to the final point, a trajectory is required to perform the desired action. The foot trajectory is designed based on collected data on the human foot position on the vertical axis. The data is collected while walking on a treadmill with an IMU-based motion capture system, Xsens \cite{Xsens}. Based on this data fifth-order splines are generated to follow these trajectories where the peak occurs at the $40\%$ percent of the total duration with a peak foot height of $0.07m$. The comparison with collected data to the designed trajectory for a motion of $1s$ is illustrated in Fig.~\ref{fig:trajectoryDesign}.

The speed and acceleration of the start and end of each joint are taken as zero. For the in-plane motion, the initial position is taken as zero where the final step position is given by the CoP position as the result of the optimization problem.

\begin{figure}[b]
    \centering
    \includegraphics[width=1\linewidth]{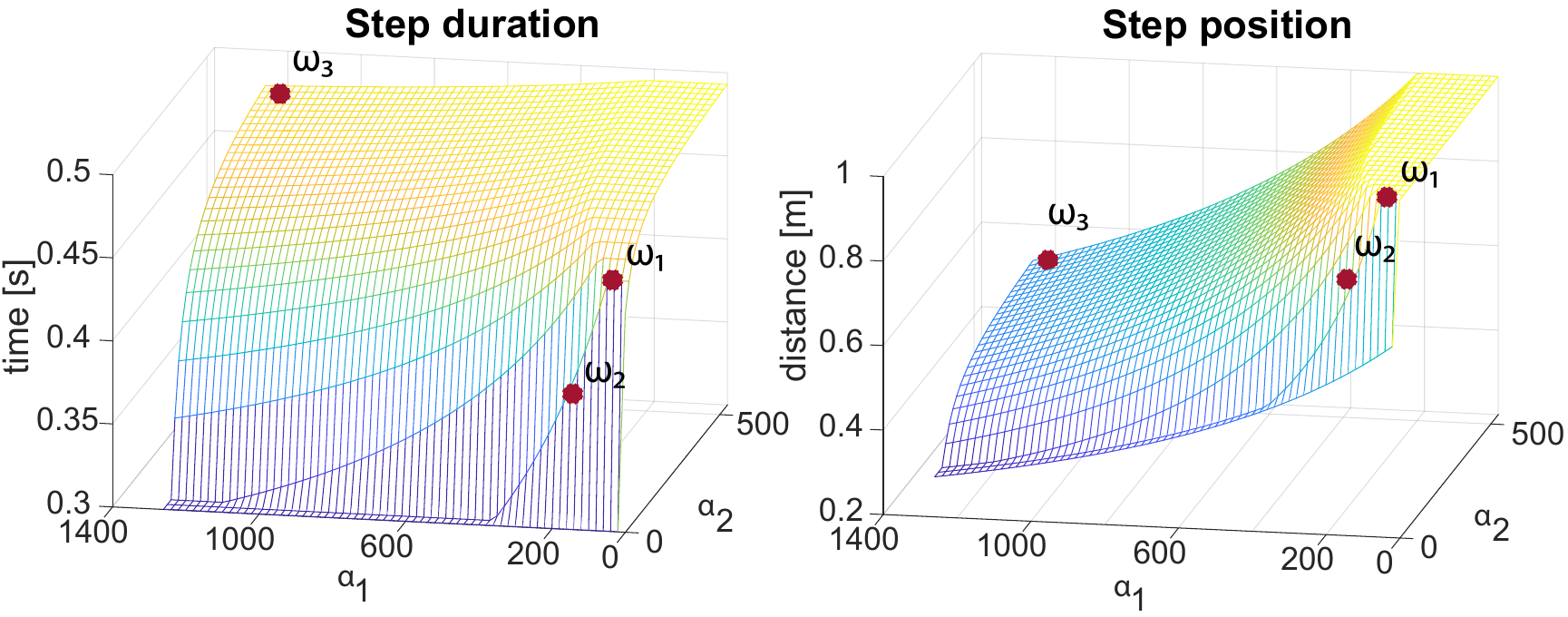}
    \vspace{-0.1cm}
\caption{The tested optimization weight vectors ($w_1$, $w_2$, $w_3$) resulted in step durations and next step positions.}
\label{fig:optWeights}
\end{figure}

\begin{figure*}[h!]
    \centering
    \includegraphics[width=0.82\linewidth]{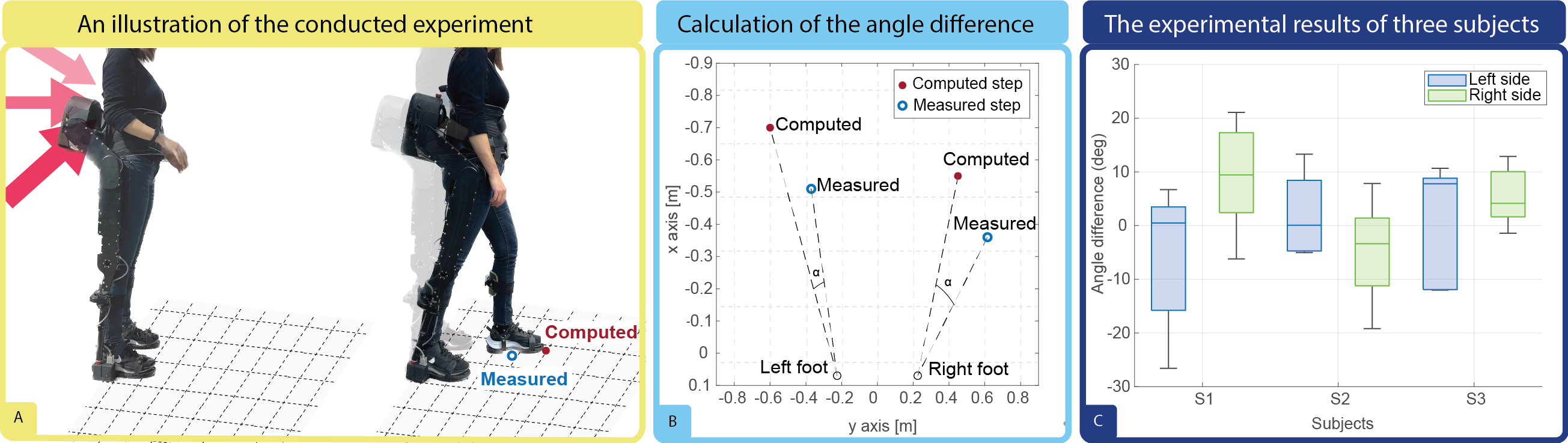}\\
    \caption{(A) An illustration of the conducted experiment where on the left subject is standing straight and on the right the subject is landed on the recovery step position. (B) Example of the angle difference between measured CoP position and the computed step positions during zero-torque control mode experiments. (C) The experimental results of three subjects show the computed angle differences between the actual recovery center of pressure positions and the computed center of pressure positions next during push recovery in zero-torque control mode when the exoskeleton was not applying assistive torques in the desired trajectory.}
    \label{fig:experimentCond}
\end{figure*}

\subsubsection{Impedance Controller}

An impedance controller strategy is selected for the implementation of recovery stepping since it provides safe and intuitive assistance during balance recovery in cooperation with the user. As suggested in \cite{Vallery2012}, open-loop assistance is provided, where the exoskeleton only assists when the balance loss is detected. To perform the recovery step, a virtual spring is rendered around the desired trajectory of the joint angles.

The controller architecture for the hip flex./ext. and knee/flex./ext. is shown in Fig.\ref{fig:controldiagram} where the impedance control loop is followed by the inner P-torque control loop. Spring stiffness of impedance controller is selected as [1.5-0.4-0.4] Nm/deg for hip abd./add., and hip and knee flex./ext., respectively. Since there is no torque sensor placed at the hip abd./add. joint, torque is controlled over the motor current.

When there is no need to take a recovery step or once the stepping is finished, the exoskeleton is in a zero-torque control mode where the spring stiffness of the impedance controller is set to zero to interfere with the movement of the subject minimally. 

\vspace{-0.2cm}
\section{Experimental Results}

\subsection{Effect of Optimization Parameters on Recovery Step}
The constrained optimization problem from Sec.~\ref{sec:stepAdaptation} is solved using the open-source software qpOASES \cite{Ferreau2014}. To be able to generate more natural recovery steps in terms of step duration and next CoP position, several distinct weight combinations are tested as $w_1$, $w_2$, and $w_3$ as shown in Fig.~\ref{fig:optWeights}. While trying to minimize the DCM offset, to be able to cooperate with the exoskeleton user, a set that resulted in shorter step lengths with longer durations, $w_3$ is selected.

\subsection{Real-World Experiments}
Three healthy subjects with the age of 27 $\pm$ 2.65 years, a height of 168.3 $\pm$  3.78 cm, and a body mass of 67.67 $\pm$  5.51 kg were asked to perform the trial. The experimental scenario included two different tests one with zero-torque mode and one with assistive mode. In the zero-torque mode, the exoskeleton can be perceived as in transparent mode where the users should ideally limited by the passive dynamics of the system. In the assistive mode, the calculated torques are provided based on the controller scheme explained in Sec.~\ref{sec:ControlStrategy}. 

During the experiments, participants stood upright on a measurement grid, facing towards the large surround. We instructed subjects to maintain a standing balance throughout each trial. The subjects are pushed from various directions from their back as depicted in Fig.\ref{fig:experimentCond}-(A). If the subject was taking a step to recover from the external push, the step is logged and recorded on the grid manually in a simultaneous manner. If the subject was using another recovery strategy the step is not recorded. Also, if the found solution was not in the direction of the external perturbation due to incoherent CoM movements by the resistance of the user, the steps are not considered.  

Although the developed recovery stepping strategy is implemented omnidirectional, it has been tested in the forward direction since fear of falling backward could potentially lead to inaccurate CoM changes \cite{Lippi2020}. Also, with our push strategy since the subjects were more prone to use the loading-unloading balance strategy side steps were not tested.

\subsubsection{Experiments in Zero-torque Mode}

Our push recovery experiments were conducted under zero-torque control mode, with the aim of evaluating the intuitiveness of the derived trajectories and analyzing the extent of symbiotic interaction between the exoskeleton and the user's intentions. For each subject 5 recovery steps are taken into account per each side as left and right. During these recovery steps in response to perturbations, we compared the measured step positions with the computed step positions, employing angle differences as a primary metric. An illustrative example of this comparison is presented in Fig.~\ref{fig:experimentCond}-(B), with the swing leg serving as the reference point for the drawn lines. The results of this analysis are presented in Fig.~\ref{fig:experimentCond}-(C).

\begin{figure}[t!]
    \centering
    \includegraphics[width=0.90\linewidth]{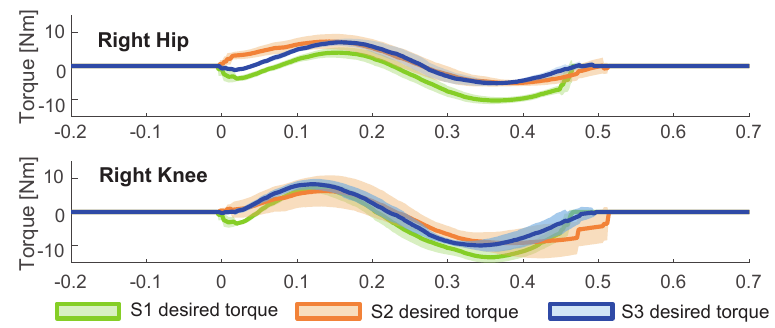}
\caption{The desired torques are depicted for right hip flex./ext. and knee flex./ext joints after the impedance controller during the experiments}
\label{fig:desiredTorquesRightK04}
\end{figure}

\subsubsection{Experiments with the Controller}

\begin{figure*}[t!]
    \centering
    \includegraphics[width=0.92\linewidth]{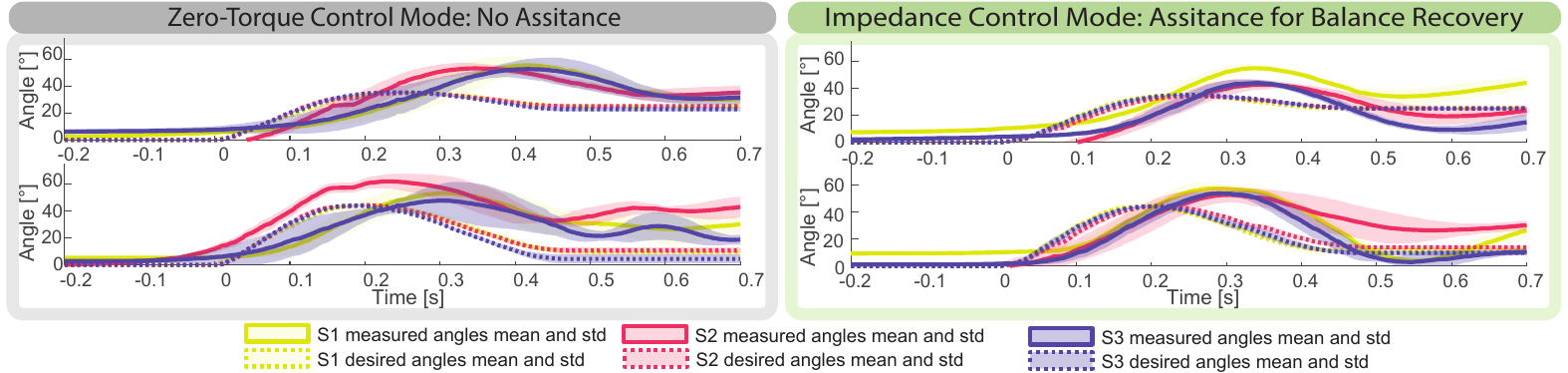}
\caption{The experimental results are presented with joint trajectories where the desired positions are computed via inverse kinematics of the found step position and measured trajectories are from exoskeleton joints. Joint trajectories for zero-torque control mode and impedance controller are given for the right side.}
\label{fig:jointTrajectoriesZTandK04}
\end{figure*}

We present the joint trajectories, in both the zero-torque control mode and the balance recovery assistance achieved through the impedance controller in Fig.~\ref{fig:jointTrajectoriesZTandK04}. The desired torques for hip flex./ext. and knee flex./ext joints are depicted for the right side as an example of the contribution of the exoskeleton in Fig.~\ref{fig:desiredTorquesRightK04}. 

\vspace{-0.3cm}
\section{Discussion} \label{Discussion}

In Fig.~\ref{fig:optWeights}, we present the outcomes of our systematic exploration of weight parameter combinations within the optimization problem. These experiments highlight our capacity to modulate step durations and step lengths effectively. For the real-world experiments, we selected a specific weight configuration, denoted as $w_3$, which yielded a stepping pattern characterized by shorter step lengths and longer durations compared to other weight settings. It's worth emphasizing that these weight values are not rigidly predetermined. Instead, they should be tuned to individual users' needs, physical conditions, and personal preferences. 

The results of the initial set of experiments conducted under zero-torque control are presented in Fig.~\ref{fig:experimentCond}-(C), showing that the mean difference in the angle of landed position to computed recovery position is in between [$-5\degree - 10\degree$]. It's important to acknowledge that that the computed position is based on the simplified kinematic model of the exoskeleton and simple CoM position estimation while the measured position is based on real-world experimental data. Given the inherent inaccuracies in the modeling and CoM estimation, this comparison may not provide a fair assessment of the computed position's accuracy. Nevertheless, these results provide valuable insights into the effectiveness of our approach in aligning the exoskeleton's contribution with user intentions. 

For the same experiment the desired and measured joint trajectories as presented in Fig.~\ref{fig:jointTrajectoriesZTandK04}, reveal our success in achieving this objective, as evidenced by the shape of the trajectories that closely resemble those of human trajectories. These findings underscore the potential of our approach to facilitate intuitive and natural interactions between users and the exoskeleton. Note that certain deviations exist in the maximum values and peak timing of hip and knee joint flex./ext. when compared to the computed trajectories. Consequently, these results suggest improvement in the design of foot trajectories, ensuring their alignment with the specific dynamics of recovery steps, rather than relying on walking trajectories as employed in this study.

For the experiments, in assistance mode our results show that with the exoskeleton's assistance, the trajectories become notably smoother and exhibit reduced oscillations with a reduction in the standard deviation of the trajectory as shown in Fig.~\ref{fig:jointTrajectoriesZTandK04}. The application of assistive torques depicted in Fig.~\ref{fig:desiredTorquesRightK04} highlights the exoskeleton's active contribution to trajectory control. These improvements show the exoskeleton's ability to enhance a more seamless recovery stepping.

Meanwhile, it should also be noted that since in the suggested approach assistance is only applied during stepping, there is a steady-state error in the final pose. Also, due to the nature of the impedance controller, the user is allowed to deviate from the desired trajectory. The controller gain should be adaptive based on the need for the assistance of the user to be able to supplement the required assistance.

In summary,  real-world experiments present the potential of our approach to align exoskeleton behavior with user intentions, enhance trajectory precision, and contribute to more synergistic recovery stepping. Nevertheless, ongoing research and refinement are essential to address identified limitations and further optimize the integration of exoskeleton assistance for various users and conditions.

\vspace{-0.2cm}
\section{Conclusion} \label{Conclusion}
\vspace{-0.1cm}

In this study, we introduce a push recovery framework for collaborative human-exoskeleton systems, which relies on a stepping strategy. We formulate the stepping strategy as an online optimization problem aimed at determining the optimal step position and duration, allowing for the recovery of balance under severe external disturbances. Our results demonstrate that the proposed framework consistently generates omnidirectional stride parameters that align closely with those of a human subject. We believe that this paper represents a useful starting point for implementing collaborative push recovery of synergistic cooperation between exoskeletons and humans.

Future endeavors should focus on the assessment of user effort and explore conditions conducive to the application of reactive stepping during various daily living activities. Moreover, systematic evaluations involving healthy control groups and individuals with balance deficiencies are warranted to refine our approach. Incorporating hip and ankle strategies in subsequent studies holds the potential to improve the versatility of the approach. Apart from its potential for standing balance, the suggested strategy could be combined with balance assistance during gait. Furthermore, Incorporating deep reinforcement learning and central pattern generators into the current framework can enhance push recovery during locomotion and promote consistent behavior across various scenarios and individuals~\cite{shafiee2023deeptransition,shafiee2023manyquadrupeds,mehr2023deep}.

\balance

\bibliographystyle{IEEEtran}
\bibliography{bibliography.bib}


\end{document}